\documentclass[final,5p,times,twocolumn]{elsarticle}
\usepackage{times}
\usepackage{epsfig}
\usepackage{graphicx}
\usepackage{amsmath}
\usepackage{amssymb}
\usepackage{epsfig}
\usepackage{commath}
\usepackage{flexisym}
\usepackage{multirow}
\usepackage[table,xcdraw]{xcolor}
\usepackage{caption}
\usepackage{subcaption}
\usepackage{comment}
\usepackage{fancyhdr}
\usepackage{scrextend}%for adding margin in algorithm
\usepackage{setspace}
\usepackage{algorithm}% http://ctan.org/pkg/algorithms
\usepackage{algpseudocode}% http://ctan.org/pkg/algorithmicx
\usepackage[pagebackref=true,breaklinks=true,colorlinks,bookmarks=false]{hyperref}
\usepackage{algorithm} 
\usepackage{algpseudocode}

\begin{document}
\begin{frontmatter}

\pagestyle{fancy}

\title{Joint Triplet Autoencoder for Histopathological Colon Cancer Nuclei Retrieval}

\author{Satya Rajendra Singh, Shiv Ram Dubey, Shruthi MS, Sairathan Ventrapragada, and Saivamshi Salla Dasharatha }
\address{
Computer Vision Group, Indian Institute of Information Technology, Sri City, Chittoor, Andhra Pradesh- 517646, India.
\\[\bigskipamount]
{\texttt{satyarajendra.rs@iiits.in, shivram1987@gmail.com, shruthi.ms16@iiits.in, sairathan.v16@iiits.in, saivamshi.s16@iiits.in}}}

% \thanks{This research is funded by Science and Engineering Research Board (SERB), Govt. of India through Project Sanction Number ECR/2017/000082. The authors would like to thank NVIDIA Corporation for the support of 2 GeForce Titan X Pascal GPU.}

\begin{abstract}
Deep learning has shown a great improvement in the performance of visual tasks. Image retrieval is the task of extracting the visually similar images from a database for a query image. The feature matching is performed to rank the images. Various hand-designed features have been derived in past to represent the images. Nowadays, the power of deep learning is being utilized for automatic feature learning from data in the field of biomedical image analysis. Autoencoder and Siamese networks are two deep learning models to learn the latent space (i.e., features or embedding). Autoencoder works based on the reconstruction of the image from latent space. Siamese network utilizes the triplets to learn the intra-class similarity and inter-class dissimilarity. Moreover, Autoencoder is unsupervised, whereas Siamese network is supervised. We propose a Joint Triplet Autoencoder Network (JTANet) by facilitating the triplet learning in autoencoder framework. A joint supervised learning for Siamese network and unsupervised learning for Autoencoder is performed. Moreover, the Encoder network of Autoencoder is shared with Siamese network and referred as the Siamcoder network. The features are extracted by using the trained Siamcoder network for retrieval purpose. The experiments are performed over Histopathological Routine Colon Cancer dataset. We have observed the promising performance using the proposed JTANet model against the Autoencoder and Siamese models for colon cancer nuclei retrieval in histopathological images.
\end{abstract}

% \begin{keywords}
% Retrieval, Siamese, Autoencoder, Deep Learning, Colon Cancer.
% \end{kywords}

\end{frontmatter}

\section{Introduction}
Image retrieval is one of the important problems of computer vision to retrieve the visually matching images from a dataset for a given query image \cite{liu2007survey}.
% , \cite{aggarwal2019new}. 
The ranking of the images is generally carried out by matching the features of a query image with the features of dataset images. Thus, the performance of retrieval depends upon the quality of the features extracted from the images which should be discriminative, robust and low dimensional \cite{pietikainen2011computer}. Several hand-designed features have been explored in the recent past, such as Local Binary Pattern (LBP) \cite{lbp}, Local Tetra Pattern (LTrP) \cite{ltrp}, Multichannel Decoded LBP (mdLBP) \cite{mdlbp}, Scale Invariant Feature Transform (SIFT) \cite{sift}, Interleaved Order-based Local Descriptor (IOLD) \cite{iold}, etc. The hand-designed features have been also explored for biomedical image retrieval such as 
% Zernike moments \cite{kumar2018efficient}, 
Local Wavelet Pattern (LWP) \cite{lwp}, Local Mesh Patterns (LMeP) \cite{lmep}, Local Bit-plane Decoded Pattern (LBDP) \cite{lbdp}, Local Ternary Co-occurrence Patterns (LTCoP) \cite{ltcop}, Local Diagonal Extrema Pattern (LDEP) \cite{ldep}, 
% Multi-scale and Multichannel Decoder based LTP \cite{sukhia2019content}, 
% Histogram of Compressed Scattering Coefficients (HCSC) \cite{lan2016medical}, Latent Dirichlet Allocation (LDA) \cite{ma2016breast}, 
etc. 

% Zheng et al. have proposed the size-scalable features for retrieval from whole slide histopathological images \cite{zheng2017size}. Lei et al. have used adaptive ensemble manifold learning over the features for neuroimaging retrieval \cite{lei2018neuroimaging}. 
% A learning to rank method \cite{xu2016improve} and a learning to refine expansion method \cite{xu2018learning} are utilized for biomedical image retrieval in Genomics datasets. However, the performance of such hand-engineered mathods is limited due to the lack of representation of highly discriminative and abstract features.

\begin{figure*}[!h]
    \centering
    \includegraphics[width=\linewidth]{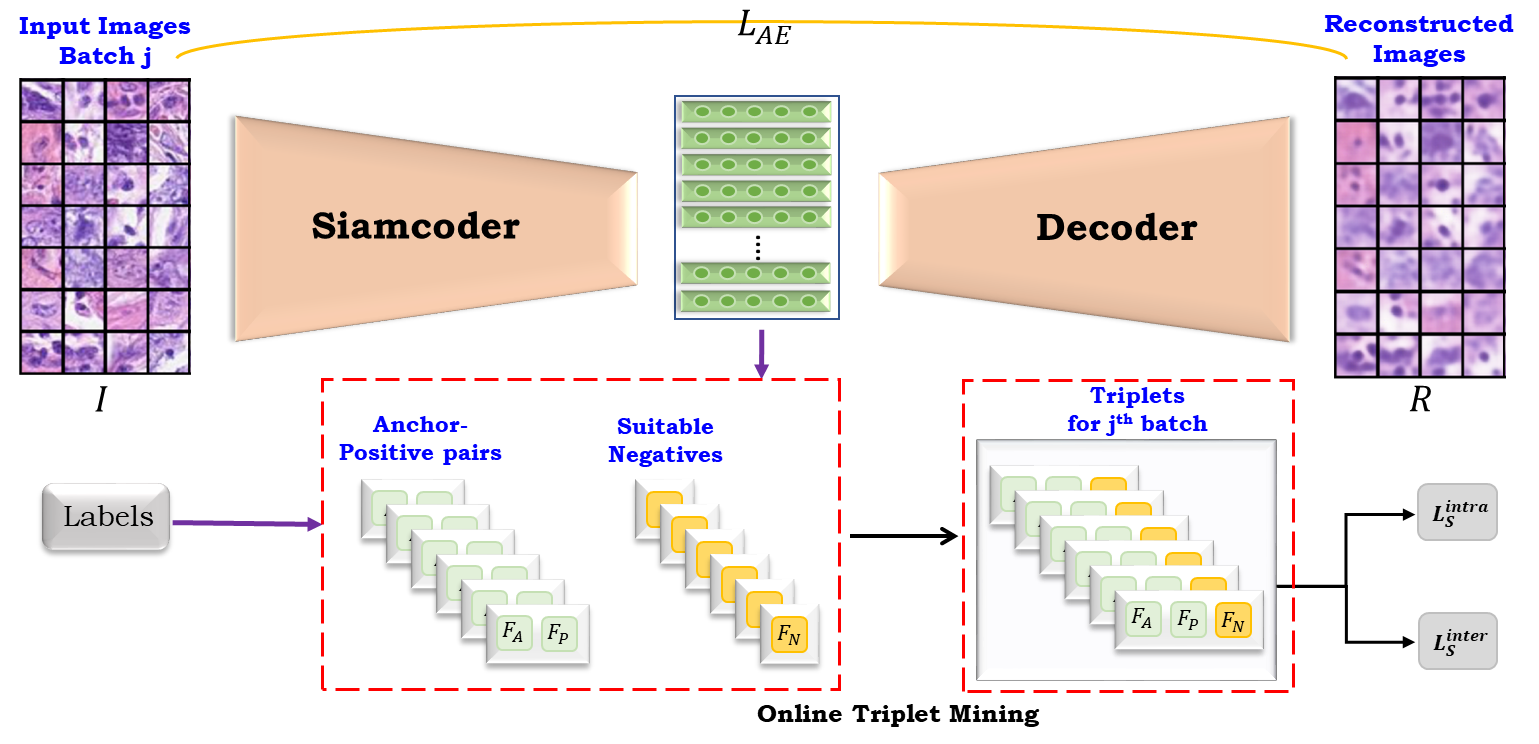}
    \caption{The proposed Joint Triplet Autoencoder Network (JTANet) model for feature learning using Autoencoder and Siamese networks in joint fashion using triplets generated through online triplet mining.}
    \label{fig:JTA_model}
\end{figure*}

In the recent past, a paradigm shift has been observed from the hand-designed feature extraction to the data driven feature learning. Thanks to the recently emerged Deep learning \cite{lecun2015deep} which facilitates the feature learning automatically from the data. Convolutional Neural Network (CNN) is a type of Neural Network designed to deal with the image and video data. AlexNet was the first CNN model developed for image classification problem in 2012. The imagenet visual recognition challenge was won by AlexNet in 2012 with a great margin as compared to the best performing hand-designed features. Since 2012, various CNN models have been proposed for different applications such as image classification 
% \cite{googlenet}, 
\cite{resnet}, 
object detection \cite{fasterrcnn}, 
% \cite{ssd}, 
image segmentation 
\cite{maskrcnn}, 
% \cite{masksrcnn}, 
% \cite{zhao2021automated},
% \cite{byra2020breast}, 
face recognition/retrieval 
% \cite{deepface}, 
\cite{srivastava2019hard}, 
% \cite{ozturk2021class}, 
% \cite{facenet}, 
\cite{dubey2018average}, 
face anti-spoofing \cite{li2018learning}, \cite{nagpal2019performance}, facial micro-expression recognition \cite{patel2016selective}, \cite{reddy2019spontaneous}, hyperspectral image classification \cite{zhao2016spectral}, \cite{hybridsn}, image-to-image translation \cite{isola2017image}, \cite{csgan}, \cite{cdgan}, and many more. 

% The analysis of histopathological images is of utmost priority for the detection, classification and retrieval of colon cancer nuclei patches \cite{rathore2013recent}. 
% An enormous growth has been also observed in deep learning area to solve the medical problems \cite{sevakula2018transfer}. 
Sirinukunwattana et al. have introduced a Spatially Constrained Convolutional Neural Network (SC-CNN) for histopathological routine colon cancer (RCC) nuclei detection and recognition \cite{sirinukunwattana2016locality}. Moreover, they have also collected the RCC nuclei dataset which is used in this paper for the experiments in retrieval framework. Recently, Basha et al. have developed a RCCNet CNN model having less number of parameters for the classification of RCC Nuclei patches \cite{rccnet}. Rajpurkar et al. have developed a ChexNet CNN model over chest X-rays data for pneumonia detection \cite{rajpurkar2017chexnet}. Wang et al. have proposed a Text-image embedding network (TieNet) for recognising the thorax disease in chest X-rays \cite{wang2018tienet}. Recently, the deep learning based methods are also proposed for biomedical image retrieval \cite{dubey2019local}, \cite{deepak2020retrieval}, \cite{baazaoui2020dynamic}. Gu and Yang have used the dense connection for multi-magnification hashing applied over histopathological images \cite{gu2018densely}.
A very recently, Sun et al. have used the adversarial learning for lesion detection \cite{sun2020adversarial}. Li et al. have developed a dual-channel deep neural network to identify the antiviral peptides \cite{li2020deepavp}. 

Autoencoder is one of the type of Neural Network which tries to learn the latent space through reconstruction process \cite{le2015tutorial}. Basically, first the input image is projected to a latent space (i.e., feature space) using an Encoder network and then it is reconstructed from that latent space using a Decoder network. The loss between the original image and the reconstructed image is minimized using Stochastic Gradient Descent (SGD) Optimization to learn the Encoder and the Decoder networks. Krizhevsky and Hinton have used a deep autoencoder for content-based image retrieval \cite{krizhevsky2011using}. Zhang et al. have used the stack of sparse autoencoder and fused its features for histopathology image analysis \cite{zhang2015fusing}. Leng et al. have utilised the autoencoder with CNN for 3D object retrieval \cite{leng20153d}. Zhu et al. have also exploited the features learnt through autoencoder for 3D shape retrieval \cite{zhu2016deep}. 
% Feng et al. have used the autoconder with deep manifold preservation for breast cancer histopathological image classification \cite{feng2018deep}.
The autoencoder has been also used in medical area such as autoencoder-based hybrid CNN-LSTM model for COVID-19 severity prediction \cite{dastider2021integrated}, detection of interacting protein pairs via ensemble of autoencoder and LightGBM \cite{sharma2020ae},
Convolutional Autoencoders to study the Alzheimer's Disease \cite{martinez2019studying} and Optimizing autoencoders based network to analyze health data \cite{zhou2018optimizing}. 
% and Graph convolutional autoencoder for predicting drug-target interactions \cite{sun2020graph}.
% Stacked sparse autoencoder for breast cancer nuclei detection \cite{xu2015stacked}, zero-bias convolutional auto-encoders for medical image classification \cite{ahn2020unsupervised} and Quadratic autoencoder for CT denoising \cite{fan2019quadratic}. 
The major drawback of such models is that they are completely unsupervised and not able to learn the discriminative features.

Siamese network is a supervised learning framework to learn the features from triplets \cite{wang2016deep}. A triplet contains three images with two from same class and one from different class. Siamese network tries to minimize the intra-class distance and maximize the inter-class distance. A pair-wise cosine loss and quantization loss is used by Cao et al. to learn the feature for image retrieval \cite{cao2016deep}. Further, they have introduced a HashNet model for retrieval \cite{cao2017hashnet}. A triplet ranking loss is used by Yao et al. for semantic preserving image retrieval \cite{yao2016deep}. A deep supervised hashing is developed by Liu et al. by utilizing the discriminative and binarization loss \cite{liu2016deep}. Yang et al. have introduced a semantic preserving deep hash (SSDH) code using CNN for image retrieval \cite{yang2017supervised}. Li et al. have developed the deep supervised discrete hashing method to learn the binary features \cite{li2017deep}. Zhang et al. have introduced a semi-supervised hashing model by incorporating the supervised classification in semi-supervised hash learning framework \cite{zhang2017ssdh}. An asymmetric deep supervised hashing is proposed by Jiang and Li by incorporating asymmetric pairwise loss \cite{jiang2018asymmetric}. Recently, Wu et al. have proposed a deep incremental hashing network to learn the hash codes of new images without changing the hash code of existing images \cite{wu2019deep}. Siamese neural networks are also used for spinal metastasis detection \cite{wang2017multi}.
% The siamese network has been also employed for medical applications such as Multi-task siamese network for retinal artery separation \cite{wang2020multi} and Margin ranking loss based siamese network for lung nodule analysis \cite{liu2019multi}. 
The major problem with Siamese network is that the learnt feature is very specific to that dataset. Moreover, it is derived only from one-way mapping (i.e., no mapping from feature to image), thus the discriminative ability might be compromized over the unseen data.

It is observed from the literature that Siamese networks and Autoencoder networks have its limitation in terms of the generalizability and discriminativeness. We propose a Joint triplet Autoencoder Network (JTANet) to learn the Encoder network of Autoencoder as a Siamese network which is termed as a Siamcoder network in this paper. Basically the triplets are used for learning of Siamcoder network in a joint fashion. The main contributions of this paper are as follows:
\begin{itemize}
    \item A semi-supervised Joint Triplet Autoencoder Network (JTANet) is proposed by utilizing the Siamese and Autoencoder networks.
    \item A Siamcoder network is used as a common CNN for Siamese network as well as Encoder network of Autoencoder.
    \item A joint training is performed for supervised Siamese network and unsupervised Autoencoder network.
    \item The latent space output of Siamcoder network is used as the feature vector/embedding for the retrieval purpose.
    \item The joint training enhances the generalizability and discriminativeness of the latent space.
    \item The histopathological colon cancer nuclei retrieval experiments are performed by using the features derived from the learnt Siamcoder network of the JTANet model.
    \item The effect of different losses is analyzed through experiments.
\end{itemize}

The organization of the paper is as follows. Section II describes the proposed JTANet model; Section III presents the histopathological colon cancer nuclei retrieval framework using JTANet model; Section IV describes the experimental settings; Section V illustrates the experimental results and analysis; and finally, Section VI concludes the paper.
% with summarizing remarks.

\begin{figure}[!t]
\centerline{\includegraphics[width=0.9\linewidth]{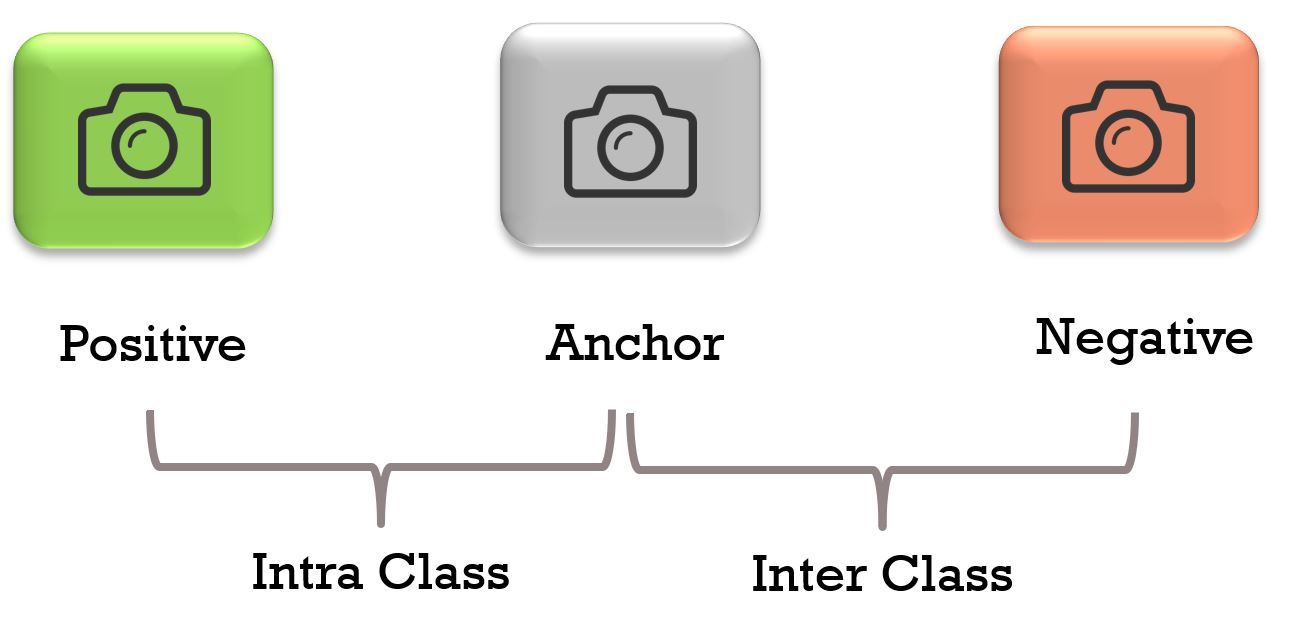}}
\caption{The triplet depicting the intra-class and inter-class samples.}
\label{fig:triplet}
\end{figure}

\section{Proposed Joint Triplet Autoencoder Network}
A Joint Triplet Autoencoder Network (JTANet) is proposed in this paper which combines the power of Siamese and Autoencoder networks. The proposed JTANet model is illustrated in Fig. \ref{fig:JTA_model}. A Siamcoder network ($SCN$) (basically the encoder network of Autoencoder) is used to transform an image patch ($I$) into the feature/latent space/embedding ($F$). Note that the Siamcoder network is shared between the Siamese network ($SN$) and Autoencoder network ($AN$). The Siamcoder network is a Convolutional Neural Network (CNN) having different layers such as convolution, batch normalization, activation function, and max pooling. Basically, the non-linear transformation function of Siamcoder network is denoted as $f_{SCN}: \mathbb{R}^{m \times m \times 3} \rightarrow \mathbb{R}^{dim}$ which converts the two-dimensional data of size $m \times m$ having three channels into an one-dimensional feature vector of size $dim$. Thus, $f_{SCN}$ can be seen as a hierarchy of some linear and non-linear functions. The Decoder network ($DN$) of Autoencoder is an Up-Convolutional Neural Network to reconstruct the output image from the latent space which is corresponding to the original image. It is also a non-linear function having different layers such as transpose-convolution, batch normalization, activation function, and upsampling. The transformation function for the Decoder network is denoted as $f_{DN}: \mathbb{R}^{dim} \rightarrow \mathbb{R}^{m \times m \times 3}$ which transforms the one-dimensional feature $F$ of size $dim$ into an image patch $R$ of size $m \times m$ with three channels. Note that $R$ is the reconstructed image w.r.t. the original image $I$.

Consider $\{I_A, I_P, I_N\}$ as a triplet of patches with $C_{I_A} = C_{I_P}$ and $C_{I_A} \neq C_{I_N}$ where $C_{I_k}$ represents the class label for image patch $I_k$ for $k \in \{A,P,N\}$. The image patches $I_A$, $I_P$, and $I_N$ are referred as the Anchor, Positive and Negative samples, respectively. A pictorial representation of triplet is shown in Fig. \ref{fig:triplet}. The proposed JTANet network generates the triplets of image patches for a batch of input from its latent space using online triplet mining as portrayed in Fig. \ref{fig:JTA_model} at the training time. However, it only needs a single image at test time to extract the features. Thus, the features used by online triplet mining for a batch of images are computed by the same Siamcoder network. The feature vectors are normalized before online triplet mining. 

\subsection{Objective Function}
Three losses, namely Autoencoder loss, Siamese loss and Feature Regularization loss are used as the objective function to train the JTANet model. Mathematically, the objective function of the proposed JTANet model is given as,
\begin{equation}
    L_{JTA} = \lambda_{AE} L_{AE} + \lambda_{SM} L_{SM} + \lambda_{FR} L_{FR}
\end{equation}
where $L_{JTA}$ is the final loss function, $L_{AE}$ is the Autoencoder loss, $L_{SM}$ is the Siamese loss, $L_{FR}$ is the Feature Regularization loss and $\{\lambda_{AE}, \lambda_{SM}, \lambda_{FR}\}$ are the hyper-parameters as the weights for the Autoencoder, Siamese, and Feature Regularization losses, respectively.

\subsubsection{Autoencoder Loss}
The Autoencoder loss ($L_{AE}$) is computed between the original images $I_k$ and its corresponding regenerated images $R_k$ for a batch of input. The purpose of Autoencoder loss is to make sure that the relevant features are being learnt by Siamcoder network. It is ensured by reconstructing the image from feature/latent space. Thus, it makes sure that Siamcoder network should not learn the random features. Mathematically, $L_{AE}$ is given as,
\begin{equation}
    L_{AE} = \sum_{i=1}^{B}L_{AE_i}
\end{equation}
where $B$ is the batch size (i.e., the number of samples in a batch) and $L_{AE_i}$ is the Autoencoder loss between the $i^{th}$ input image patch $I_i$ and its corresponding reconstructed image patch $R_i$ for a batch. The $L_{AE_i}$ is computed as the mean square error (MSE) between $I_i$ and $R_i$ and given as,
\begin{equation}
    \begin{split}
    L_{AE_i} 
    & = ||I_i - R_i||^2_{avg} \\
    & = \frac{1}{m \times m \times 3} \sum_{u=1}^{m}\sum_{v=1}^{m}\sum_{c=1}^{3} (I_i(u,v,c) - R_i(u,v,c))^2        
    \end{split}
\end{equation}
where $I_i(u,v,c)$ and $R_i(u,v,c)$ denote the values in the image patches $I_i$ and $R_i$, respectively at image co-ordinate $(u,v)$ in $c^{th}$ channel and $||.||^2_{avg}$ represents the mean square error (MSE).

\begin{figure*}[!t]
    \centering
    \includegraphics[width=0.8\linewidth]{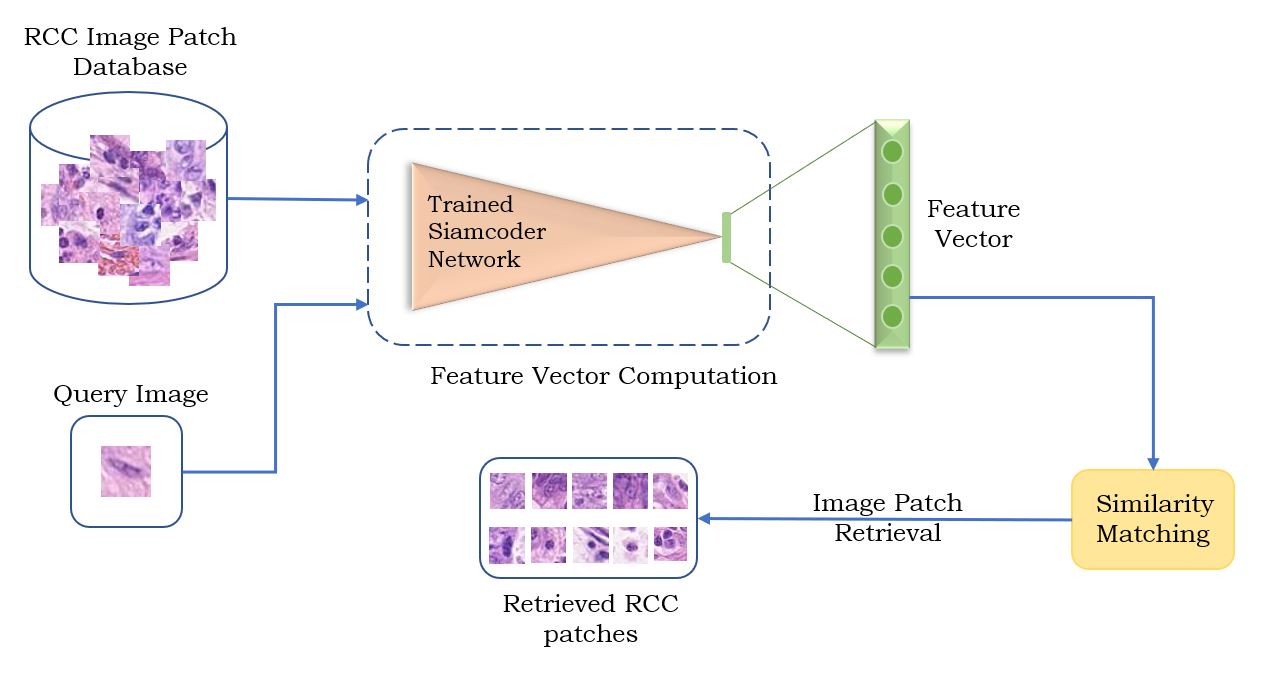}
    \caption{The image retrieval framework in Routine Colon Cancer (RCC) image patch database using the proposed trained Siamcoder network.}
    \label{fig:JTA_framework}
\end{figure*}

\subsubsection{Siamese Loss}
The online triplet mining uses the features derived from Siamcoder network to generate the triplets of the Anchor, Positive and Negative samples. Consider, $\{F_A, F_P, F_N\}$ as the Anchor, Positive and Negative triplets of Siamcoder network generated features corresponding to the Anchor, Positive and Negative image patch triplet $\{I_A, I_P, I_N\}$, respectively. The Siamese loss $(L_{SM})$ is computed by using the $\{F_A, F_P, F_N\}$ features corresponding to the Anchor, Positive and Negative samples. The purpose of Siamese loss is to decrease the distance between the features of Anchor and Positive samples and to increase the distance between the features of Anchor and Negative samples. By doing so, it forces the Siamcoder network to learn the class specific features such that the features for samples of same class are closer to each other and the features for samples of different class are apart from each other. The $L_{SM}$ is given as,
\begin{equation}
L_{SM} = \begin{cases}
        0 & L_{S} \leq 0\\
        L_{S} &\text{Otherwise}
        \end{cases}
\end{equation}
where $L_{S}$ is defined as,
\begin{equation}
    L_{S} = L_{S}^{intra} - L_{S}^{inter} + \lambda_{m}
\end{equation}
where $L_{S}$ is the Siamese loss, $L_{S}^{intra}$ is the Siamese intra-class similarity loss, $L_{S}^{inter}$ is the Siamese inter-class similarity loss, and $\lambda_{m}$ is a margin hyper-parameter. The Siamese intra-class similarity loss is computed between the features of the samples of the same class (i.e., $F_A$ and $F_P$). Similarly, the Siamese inter-class similarity loss is computed between the features of the samples of the different classes (i.e., $F_A$ and $F_N$). These losses are computed as,
\begin{equation}
    L_{S}^{intra} = \sum_{i}^{nb} ||F_A^{i} - F_P^{i}||^2 = \sum_{i}^{nb}\sum_{v=1}^{dim}(F_A^i(v) - F_P^i(v))^2
\end{equation}
and 
\begin{equation}
    L_{S}^{inter} = \sum_{i}^{nb}||F_A^i - F_N^i||^2 = \sum_{i}^{nb}\sum_{v=1}^{dim}(F_A^i(v) - F_N^i(v))^2
\end{equation}
where $F_A^i$, $F_P^i$, and $F_N^i$ are the feature vectors of size $dim$ derived using Siamcoder network for $i^{th}$ triplet having Anchor ($I_A^i$), Positive ($I_P^i$), and Negative ($I_N^i$) image patches, respectively, $nb$ is the number of triplets generated by online triplet mining for a batch of input and $||.||^2$ represents the sum of square distances (SSD).

\subsubsection{Feature Regularization Loss}
The Feature Regularization loss ($L_{FR}$) is computed from the Siamcoder network's output features ($F_i|_{1 \leq i \leq B}$). The purpose of Feature Regularization loss is to increase the generalization ability of features being produced by the the Siamcoder network. It is computed as,
\begin{equation}
    L_{FR} = \sum_{i=1}^{B}L_{FR}^i
\end{equation}
where $L_{FR}^i|_{1 \leq i \leq B}$ is the Feature Regularization loss over the feature vector $F_i$ and defined as,
\begin{equation}
    L_{FR}^i = ||F_i||^2_{norm} = \sum_{v=1}^{dim}(F_i(v))^2
\end{equation}
by calculating the $L_2$-Norm of feature vector $F_i$ for $i \in [1,B]$.

We use Stochastic Gradient Descent (SGD) based optimization techniques called as Adam to train the Siamcoder and Decoder networks of the proposed JTANet model. The weights of any network in deep learning are generally trained by updating it during backpropagation of training using SGD based update rules \cite{adam}, \cite{diffgrad}. 

\begin{figure*}[!t]
\includegraphics[trim=0 200 0 50, width=\linewidth, clip]{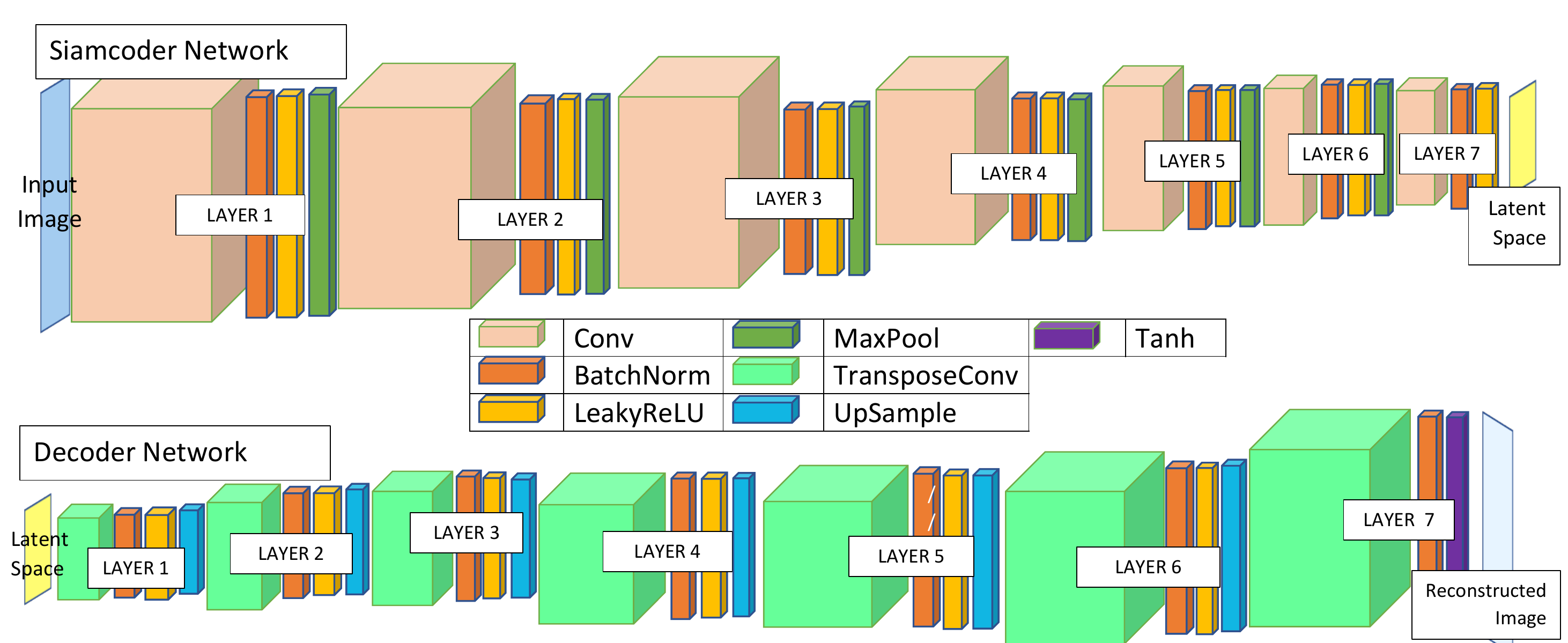}%<left> <lower> <right> <upper>
\caption{The Siamcoder and Decoder architectures of the proposed JTANet model.}
\label{fig:network}
\end{figure*}

\begin{table*}[!t]
\caption{Layer wise details of the proposed JTANet architecture. Each Conv and ConvTranspose layer uses 3x3 filters without bias with stride 1 and padding 1, each LeakyReLu layer uses 0.2 leaky factor, each MaxPool layer uses 2x2 kernal with stride 2 and each UpSample layer uses scaling factor 2 with bilinear upsampling strategy. The embedding length is represented by EL. The Tanh is the activation function in the last layer of the decoder network.}
\centering
\resizebox{\textwidth}{!}{%
\begin{tabular}{|c|l|c|c|l|c|c|}
\hline
\multirow{2}{*}{Layer}&\multicolumn{3}{c|}{SIAMCODER
Network}&\multicolumn{3}{c|}{DECODER Network}\\
\cline{2-7}
& Filter & Input Dimenstion & Output Dimenstion & Filter & Input Dimenstion & Output Dimenstion\\
\hline
1 & Conv & 64x64x3 & 64x64x64 & ConvTranspose & 1x1xEL & 1x1x1024 \\
\cline{2-7}
 & BatchNorm, LeakyReLu & 64x64x64 & 64x64x64 & BatchNorm, LeakyReLu & 1x1x1024 & 1x1x1024 \\
 \cline{2-7}
 & MaxPool & 64x64x64 & 32x32x64 & UpSample & 1x1x1024 & 2x2x1024 \\
\hline
2 & Conv & 32x32x64 & 32x32x128 & ConvTranspose & 2x2x1024 & 2x2x1024 \\
\cline{2-7}
 & BatchNorm, LeakyReLu & 32x32x128 & 32x32x128 & BatchNorm, LeakyReLu & 2x2x1024 & 2x2x1024 \\
\cline{2-7}
 & MaxPool & 32x32x128 & 16x16x128 & UpSample & 2x2x1024 & 4x4x1024 \\
\hline
3 & Conv & 16x16x128 & 16x16x256 & ConvTranspose & 4x4x1024 & 4x4x512\\
\cline{2-7}
 & BatchNorm, LeakyReLu & 16x16x256 & 16x16x256 & BatchNorm, LeakyReLu & 4x4x512 & 4x4x512\\
\cline{2-7}
 & MaxPool & 16x16x256 & 8x8x256 & UpSample & 4x4x512 & 8x8x512\\
\hline
4 & Conv & 8x8x256 & 8x8x512 & ConvTranspose & 8x8x512 & 8x8x256 \\
\cline{2-7}
 & BatchNorm, LeakyReLu & 8x8x512 & 8x8x512 & BatchNorm, LeakyReLu & 8x8x256 & 8x8x256 \\
 \cline{2-7}
 & MaxPool & 8x8x512 & 4x4x512 & UpSample & 8x8x256 & 16x16x256 \\
\hline
5 & Conv & 4x4x512 & 4x4x1024 & ConvTranspose & 16x16x256 & 16x16x128 \\
\cline{2-7}
 & BatchNorm, LeakyReLu & 4x4x1024 & 4x4x1024 & BatchNorm, LeakyReLu & 16x16x128 & 16x16x128 \\
 \cline{2-7}
 & MaxPool & 4x4x1024 & 2x2x1024 & UpSample & 16x16x128 & 32x32x128 \\
\hline
6 & Conv & 2x2x1024 & 2x2x1024 & ConvTranspose & 32x32x128 & 32x32x64 \\
\cline{2-7}
 & BatchNorm, LeakyReLu & 2x2x1024 & 2x2x1024 & BatchNorm, LeakyReLu & 32x32x64 & 32x32x64 \\
\cline{2-7}
 & MaxPool & 2x2x1024 & 1x1x1024 & UpSample & 32x32x64 & 64x64x64 \\
\hline
7 & Conv & 1x1x1024 & 1x1xEL & ConvTranspose & 64x64x64 & 64x64x3 \\
\cline{2-7}
 & BatchNorm, LeakyReLu & 1x1xEL & 1x1xEL & BatchNorm, Tanh & 64x64x3 & 64x64x3 \\
\hline
\end{tabular}}
\label{tab:jta}
\end{table*}

\section{Routine Colon Cancer Nuclei Retrieval using Proposed JTANet Model}
The proposed JTANet model is trained using the Routine Colon Cancer (RCC) Nuclei triplets. The training is performed for Siamcoder and Decoder networks in joint fashion using Autoencoder, Siamsese, and Regularization losses. Once the JTANet model is trained, only Siamcoder network is required to extract the features from any image patch as depicted in Fig. \ref{fig:JTA_framework}. A training feature database is created as follows,
\begin{equation}
    F^{train}_{i,C_i}|_{i = 1, 2, ..., N^{train}} = f_{SCN}(I^{train}_{i,C_i})
\end{equation}
where $N^{train}$ is the total number of images in the training set, $I^{train}_{i,C_i}$ is the $i^{th}$ image in the training set having $C_i$ as the class label, $f_{SCN}$ is the Siamcoder network function to transform an image patch into a feature vector, and $F^{train}_{i,C_i}$ is the extracted features for image patch $I^{train}_{i,C_i}$. Consider $I^q$ as a query image patch for which we want to retrieve the best $\delta$ number of images from training set. The features $F_q$ are also extracted for input query $I^q$ using the same trained Siamcoder network to facilitate the feature matching as,
\begin{equation}
    F^{q} = f_{SCN}(I^{q}).
\end{equation}
Note that for the experimentation purpose, the query image patch $I^q$ is taken from the test set of the Routine Colon Cancer Nuclei database. Thus, $q = 1, 2, 3, ..., N^{test}$ where $N^{test}$ is the number of images in the test set. The distance between features of query image $F^{q}$ and training images $F^{train}_{i,C_i}$ for $i = 1, 2, ..., N^{train}$ is computed to retrieve the best $\delta$ images from training set based on the distance ranking in increasing order as illustrated in Fig. \ref{fig:JTA_framework}. The Euclidean distance is computed between the features as follows:
\begin{equation}
    dis(F^{q},F^{train}_{i,C_i}) = (\sum_{z=1}^{dim}(F^{q}(z)-F^{train}_{i,C_i}(z))^2)^{0.5}.
\end{equation}

The performance of the model is computed by considering each image patch in test as the query image one by one. Consider $I^{test}_{j,C_j}|_{j = 1, 2, ..., N^{test}}$ as the $j^{th}$ sample in the test set as the query patch $q$ from $C_j^{th}$ class where $N^{test}$ is the total number of RCC image patches in the test set. The feature extracted using trained Siamcoder network for test image patch $I^{test}_{j,C_j}$ is $F^{test}_{j,C_j}$. The mean precision is computed as,
\begin{equation}
    Pr = \frac{\sum_{j=1}^{N^{test}}Pr_j}{N^{test}}
\end{equation}
where $Pr_j$ is the precision when $j^{th}$ test sample is considered as the query and computed as,
\begin{equation}
    Pr_j = 100 \times \frac{\#CR_j}{\#TR_j}
\end{equation}
where $\#CR_j$ and $\#TR_j$ are the number of correctly retrieved and total retrieved images for a query.

\section{Experimental Settings}
This section is devoted for the acrchitecture details, dataset description, triplet generation and hyper-parameter settings.

\subsection{Architecture for Siamcoder and Decoder Networks}
The proposed model uses Siamcoder and Decoder networks for feature extraction and image patch reconstruction, respectively as depicted in Fig. \ref{fig:network}. 
The Siamcoder network consists of seven convolution layers. The Decoder network also consists of seven deconvolutional layers. The layer-wise details of the proposed JTANet model is summarized in Table \ref{tab:jta}.

\begin{figure}[!t]
    \centering
    \includegraphics[width=\linewidth]{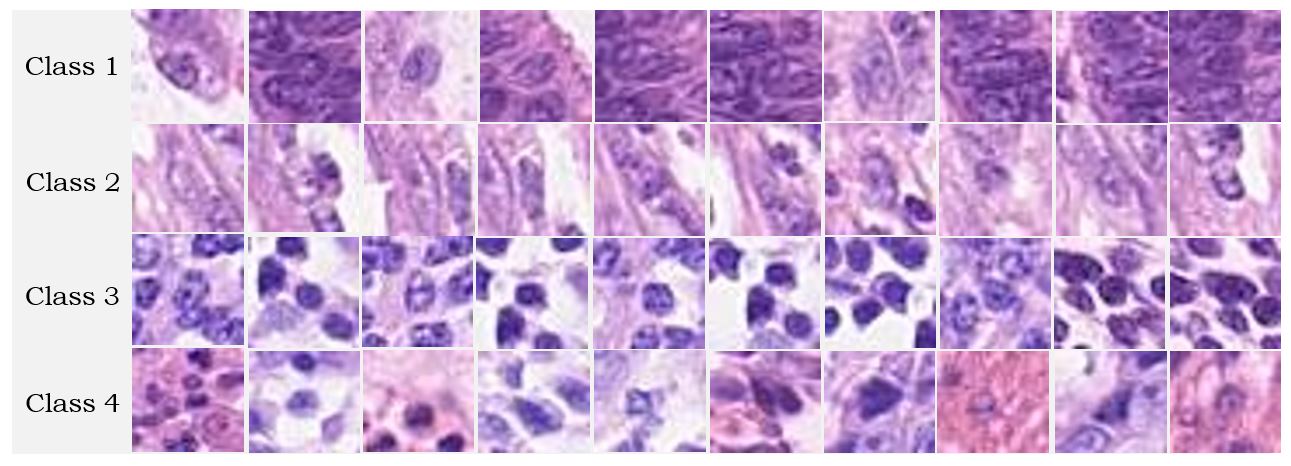}
    \caption{Sample image patches from Histopathological Routine Colon Cancer (RCC) Nuclei Cell dataset.}
    \label{fig:samples_images}
\end{figure}

\begin{figure}[!t]
\includegraphics[width=\columnwidth]{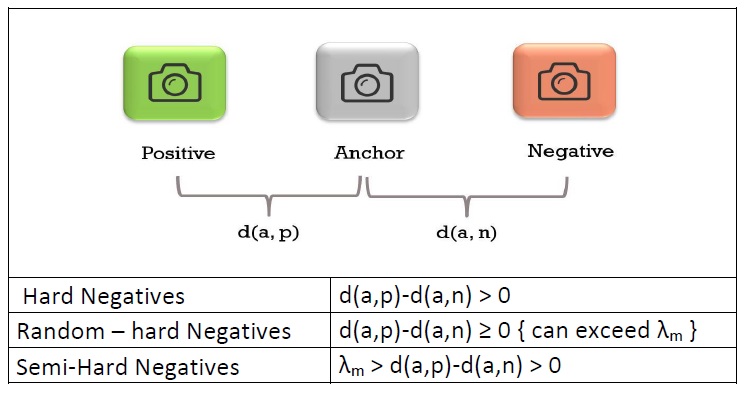}
\caption{Different strategies used for the negatives selection of triplets.}
\label{fig:anch-dia}
\end{figure}

\subsection{Routine Colon Cancer Dataset}
A CRCHistoPhenotypes - Labeled Routine Colon Cancer (RCC) Cell Nuclei Dataset\footnote{https://warwick.ac.uk/fac/sci/dcs/research/tia/data/crchistolabelednucleihe/} is used in this paper for the experiments \cite{sirinukunwattana2016locality}. This dataset contains $100$ H\&E stained histology images of colorectal adenocarcinomas. The co-ordinates of 22,444 RCC nuclei is also provided with this dataset along with its associated labels. Basically, this RCC dataset has $4$ classes which are named as, epithelial, fibroblast, inflammatory and others. We use the co-ordinates of nuclei cells to extract the $32 \times 32$ patches around the cells. The image patches are distributed unevenly among four classes (i.e., epithelial-$7,722$ patches, fibroblast-$5,712$ patches, inflammatory-$6,971$ patches, others-$2,039$ pactches). The dimension of the RCC image patches is $32 \times 32$ having $3$ channels. The sample image patches of RCC dataset is shown in Fig. \ref{fig:samples_images}. The dataset is split into train and test sets having $20,444$ and $2,000$ image patches, respectively.

\begin{figure*}[!t]
\centering
\begin{subfigure}{.33\textwidth}
\includegraphics[width=0.985\columnwidth]{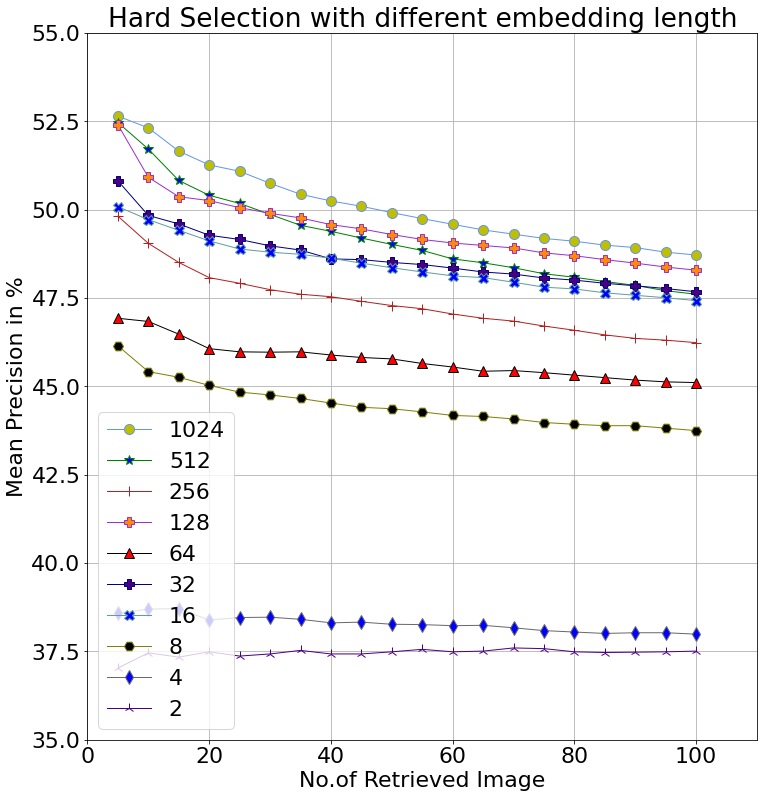}
\caption{}
\label{fig:precision_hard}
\end{subfigure}%
\begin{subfigure}{.33\textwidth}
\includegraphics[width=0.973\columnwidth]{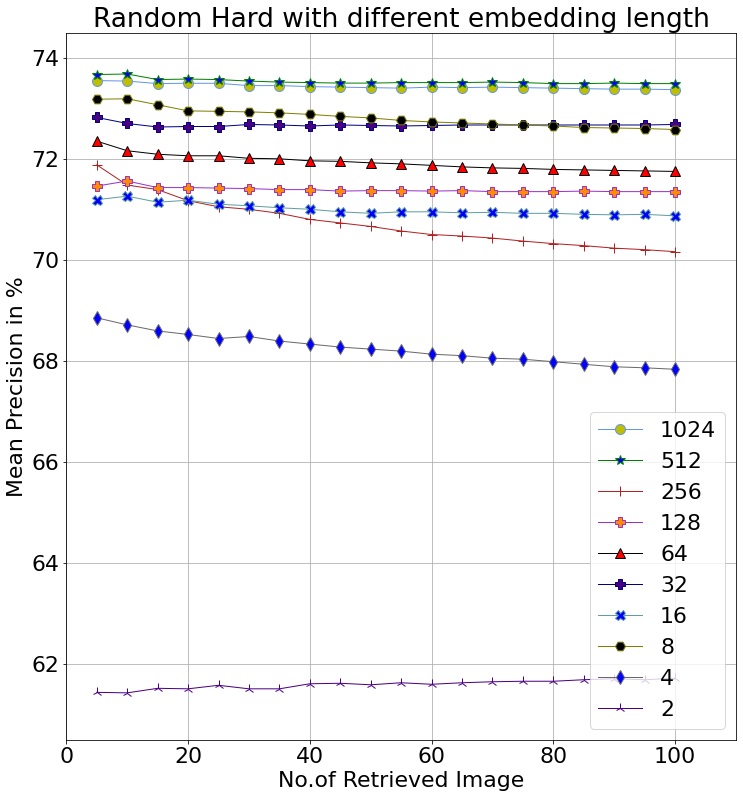}
\caption{}
\label{fig:precision_rand-hard}
\end{subfigure}%
\begin{subfigure}{.33\textwidth}
\includegraphics[width=0.974\columnwidth]{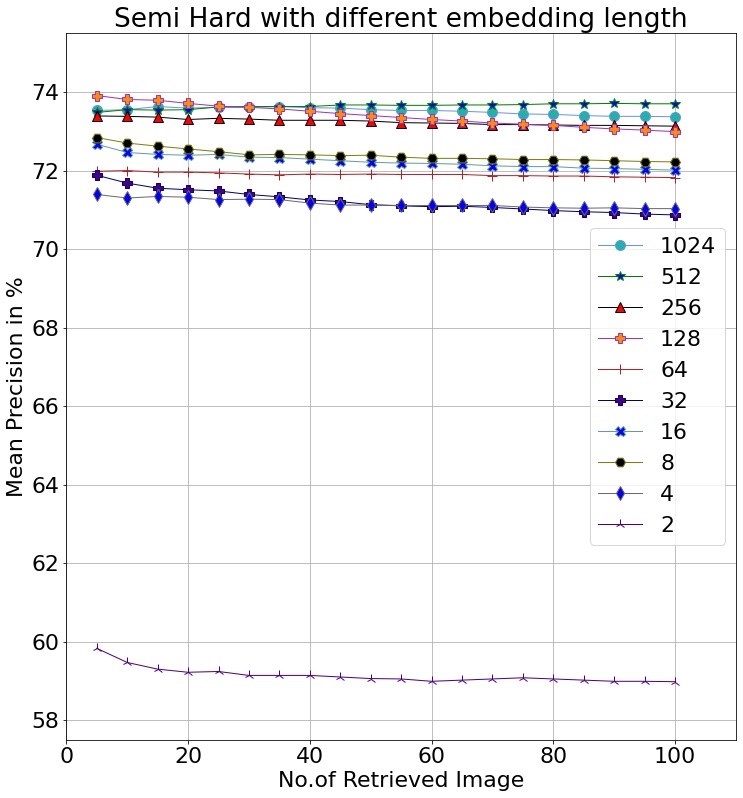}
\caption{}
\label{fig:precision_semi-hard}
\end{subfigure}%
\caption{The mean precision computed using the proposed JTANet model for different no. of retrieved images using (a) Hard, (b) Random Hard, and (c) Semi Hard negative selection strategies of triplet generation.}
\label{fig:precision}
\end{figure*}

\subsection{Triplet Generation}
% Very first approach to triplet mining was the naive method or systematic random approach where we shuffle the dataset multiple times and map the triplets making sure that none of them are repeated. This naive approach, did not yield good results as most of the triplets generated were easy triplets. This approach was implemented on CIFAR-10 dataset. An example of the triplet generated using naive method is shown in fig3.
% Since the previous method did not yield satisfying triplets, 
Online triplet mining \cite{facenet} is used in this work to generate the triplets. Only training set is used to generate the triplets as it is needed only at training time. First, the Anchor-Positive pairs are generated using the distance matrix computed from the embeddings extracted by the siamcoder network. Next, a suitable negative sample for each Anchor-Positive pair is selected to form the triplet. The negative sample is taken from any other class than anchor class using the negative selection function. A triplet score is passed as the input to the negative selection function. The triplet score is computed as $max(d(A,P) - d(A,N) + \lambda_m, 0)$ where $d$ is the distance function, $A$, $P$, and $N$ are the Anchor, Positive, and Negative samples of any triplet, and the margin $\lambda_m$ is a hyper-parameter. We use three types of negative sample selection approaches as shown in Fig \ref{fig:anch-dia} and described as follows: 
\begin{itemize}
    \item \textbf{Hard Negative: } The difference between the distances should be greater than zero. The negatives sorted in descending order of the score values and the negative with largest score value is selected for an anchor-positive pair.
    \item \textbf{Semi-hard Negative: } The difference between the distances should be greater than zero and at the same time it should not exceed the margin. All such negatives satisfying the condition are considered and one of them is selected as the suitable negative for a given anchor-positive pair.
    \item \textbf{Random-hard Negative: } A combination of both hard and semi-hard triplets is considered. The negatives are randomly selected which satisfy a common condition that the difference in the distances should be greater than zero.
\end{itemize}

\begin{table*}[!t]
\caption{The mean precision obtained using the proposed JTANet method for different embedding length for 5 no. of retrieved images.}
\centering
\begin{tabular}{|c|c|c|c|}
\hline
Embedding length & HARD & RANDOM HARD & SEMI HARD\\\hline
1024&52.64&73.56&73.52\\\hline
512&52.46&73.67&73.47 \\\hline
256&49.82&71.88&73.39\\\hline
128&52.40&71.46&73.90\\\hline
64&46.93&72.36&71.98\\\hline
32&50.82&72.82&71.87\\\hline
16&50.08&71.20&72.66\\\hline
8&46.15&73.18&72.83\\\hline
4&38.58&68.85&71.38\\\hline
2&37.03&61.43&59.82\\\hline 
\end{tabular}
\label{tab:Results}
\end{table*}

\subsection{Hyper-parameter Settings}
The experiments are conducted using the PyTorch framework. The Adam optimiser is used with $0.001$ learning rate for $50$ epochs. The batch size is set to $256$. The input patch dimension is resized to $64 \times 64 \times 3$. In random-hard and semi-hard negative selections, the margin threshold value ($\lambda_{m}$) used is $0.5$. The default weight factors/coefficients for all the loss functions are used as $1$ untill or otherwise specified.

\section{Experimental Results and Analysis}
In this section, the performance of the proposed JTANet model is reported and analyzed for different settings under retrieval framework. First, we report the experimental results for Hard, Random Hard and Semi Hard triplet selection strategies. Then, we conduct the convergence analysis and finally, the loss weight coefficient analysis is done.

\subsection{Results using Different Embedding Lenghts}
The experimental results in terms of the mean retrieval precision is demonstrated in Fig. \ref{fig:precision}. The mean precision values are plotted against the number of images retrieved from 5 to 100. The embedding lengths (i.e., the latent space feature dimension) used are 2, 4, 8, 16, 32, 64, 128, 256, 512 and 1024. The plots in Fig. \ref{fig:precision_hard}, Fig. \ref{fig:precision_rand-hard}, and Fig. \ref{fig:precision_semi-hard} correspond to the Hard, Random Hard, and Semi Hard negative selection strategies of triplet generation, respectively. Generally, the mean precision decreases with increase in the number of retrieved images. 

The best performance is achieved for 1024 embedding length in case of Hard selection. Whereas, the best performance is observed for 512 embedding length in case of Random Hard and Semi Hard selections. The possible reason of such behaviour is associated with the difficulty of the triplets which is more in case of Hard selection based strategy.
It is also noticed that the mean precision is lowest for embedding length 2 across all the plots. Moreover, the perfromance is lower with smaller embedding lengths in most of the cases due to the limited discriminative power of less features. The mean precision for Randon Hard and Semi Hard strategies is better than Hard strategy. The mean precision values for 5 number of retrieved images using the proposed JTANet method is also reported in Table \ref{tab:Results} using Hard, Random Hard and Semi Hard strategies. It shows the suitability of the proposed model for routine colon cancer patch retrieval task.

\begin{figure*}[!t]
\centering
\begin{subfigure}{.33\textwidth}
\includegraphics[clip=true, trim = 15 5 35 20, width=0.98\columnwidth]{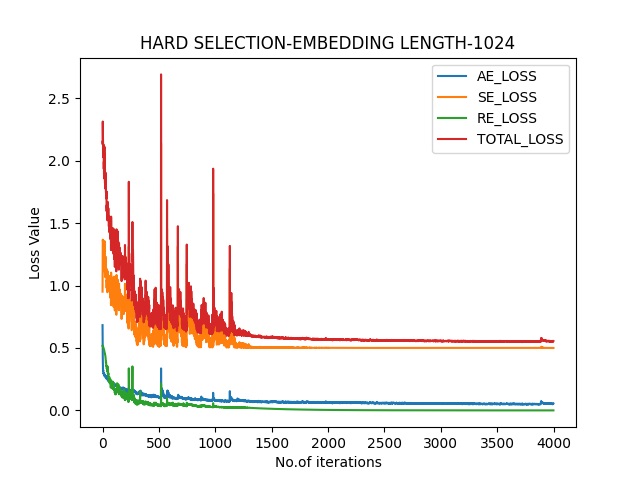}%trim - left, bottom, right, top
\caption{}
\label{fig:loss_hard}
\end{subfigure}%
\begin{subfigure}{.33\textwidth}
\includegraphics[clip=true, trim = 15 5 35 20, width=0.98\columnwidth]{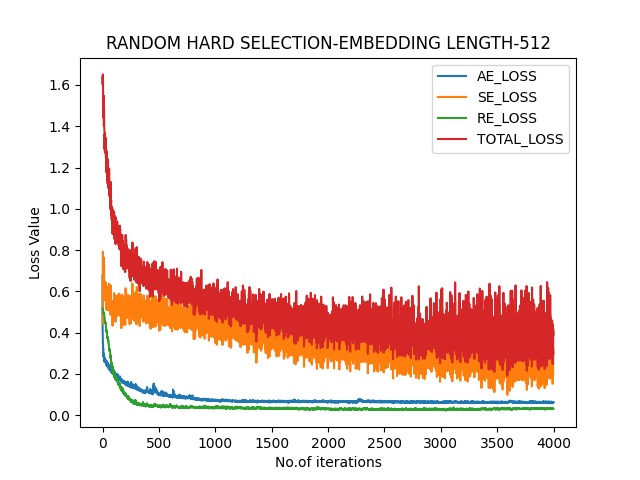}
\caption{}
\label{fig:loss_rand-hard}
\end{subfigure}%
\begin{subfigure}{.33\textwidth}
\includegraphics[clip=true, trim = 15 5 35 20, width=0.98\columnwidth]{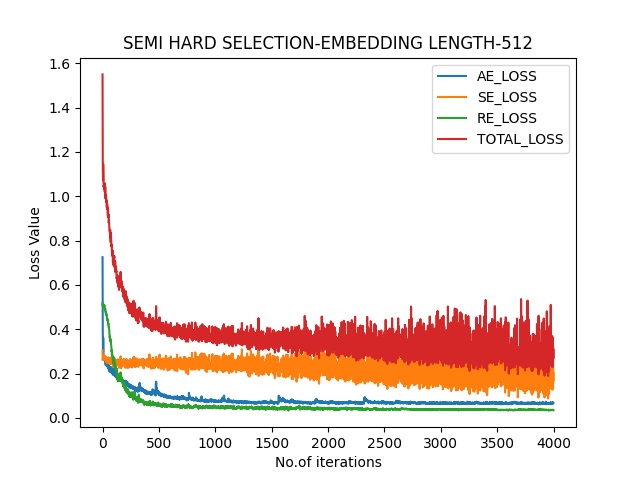}
\caption{}
\label{fig:loss_semi-hard}
\end{subfigure}%
\caption{The loss values using the proposed JTANet model w.r.t. the training iterations using (a) Hard, (b) Random Hard, and (c) Semi Hard negative selection strategies of triplet generation. Here, AE\_Loss, SE\_LOSS, RE\_LOSS, and TOTAL\_LOSS refer to autoencoder loss, siamese loss, feature regularization loss and total loss, respectively.}
\label{fig:loss}
\end{figure*}

\subsection{The Convergence Analysis}
In order to understand the training behaviour of the proposed JTANet model, the convergence analysis is done in this paper. Fig. \ref{fig:loss} shows the loss values for autoencoder loss, siamese loss, feature regularization loss, and total loss in terms of the number of the training iterations. The loss curves in Fig. \ref{fig:loss_hard}, Fig. \ref{fig:loss_rand-hard}, and Fig. \ref{fig:loss_semi-hard} correspond to the Hard, Random Hard, and Semi Hard strategies, respectively. The same unit weights are used for each loss to compute the total loss in this experiment.
The graph is plotted for Hard negative selection of triplet with the feature length 1024. Whereas, the feature length 512 is used for Random Hard and Semi Hard negative selection of triplet strategies.
It is observed from this plot that the training has been converged well w.r.t. all the loss functions. Moreover, the siamese loss dominates over other two losses. The fluctuations in the siamese loss of Random Hard as well as Semi Hard are due to the random triplets being generated in different iterations. Whereas, no randomness is present in Hard selection strategy, thus smoother loss curves have been observed. The autoencoder loss and feature regularization loss are converged within 1000 iterations in all the cases. The siamese loss also converges in reasonable number of epochs in all the cases. From this result, it is evident that the proposed JTANet model shows the faster convergence in training.

\begin{table*}[!t]
\caption{\label{tab:Results1}The mean precision in \% using the proposed JTANet model by varying the weight coefficient settings of different losses for 5 number of retrieved images. The embedding lengths used for Hard, Random-Hard and Semi-Hard triplet strategies are $1024$, $512$ and $512$, respectively.}
 \centering
\begin{tabular}{|c|c|c|c|}
\hline
\textbf{AE:SM:FR } & \textbf{Hard-1024} & \textbf{Random-Hard-512} & \textbf{Semi-Hard-512}
\\
\hline
1:1:1&52.64&73.67&73.47\\\hline
5:1:1&50.41&72.21&73.96\\\hline
1:5:1&47.90&73.50&74.18\\\hline
1:1:5&48.24 &72.03&72.85 \\\hline
10:1:1&51.14 &72.25&72.36\\\hline
1:10:1&43.65&73.43&72.86\\\hline
1:1:10&48.10&73.13&73.27\\\hline
0:1:1&58.77&74.11&73.60\\\hline
1:0:1&50.76&52.12&44.24\\\hline
1:1:0&50.09&70.78&72.23\\\hline 
\end{tabular}
\end{table*}

\subsection{The Loss Weight Coefficient Analysis}
The high precision using the proposed JTANet model is observed due the different losses used, such as autoencoder loss, siamese loss and feature regularization loss. The total loss is computed as the weighted combination of above mentioned three losses. In earlier experiments, all the weight coefficients are set as 1. In this experiment, the peformance comparison is done by varying the weight coefficients for different losses in order to compute the total loss. The mean precision (\%) using the proposed JTANet model by varying the weight settings of different losses for 5 number of retrieved images is reported in Table \ref{tab:Results1}. The feature lengths used for Hard, Random Hard and Semi Hard triplet selection strategies are 1024, 512 and 512, respectively in this experiment. 

The different weight coefficient combinations for Autoencoder Loss (AE), siamese loss (SM) and feature regularization loss (FR) (i.e., AE:SM:FR) are 1:1:1,  5:1:1, 1:5:1, 1:1:5, 10:1:1, 1:10:1, 1:1:10, 0:1:1, 1:0:1 and 1:1:0, respectively. From Table. \ref{tab:Results1}, we notice that when the autoencoder loss is missing (i.e., the weight values are 0:1:1), the mean precision is best for Hard and Random Hard triplets. However, the autoencoder loss is influential with Semi Hard triplets. 
When the weight values are 1:5:1 (i.e., more weight given to siamese loss), the mean precision is best for Semi-Hard triplets. The biased weighted highest mean precision is improved as compared to the ideal weighted (1:1:1) mean precision by 11.65\%, 0.60\%, and 0.97\% for Hard, Random Hard and Semi Hard selection strategies, respectively. 
It is also observed that very high weight for siamese loss (i.e., 1:10:1 setting) lowers the performance for Hard selection. Whereas, it is evident from 1:0:1 setting that the siamese loss is the most important loss for Random Hard and Semi Hard triplet selection strategies. We can say from this experiment that following is the relevancy of different losses: siamese loss $>$ feature regularization loss $>$ autoencoder loss in the proposed JTANet model for routine colon cancer patch retrieval.

\section{Conclusion}
This paper proposes a joint triplet autoencoder network (JTANet) for histopathological colon cancer nuclei retrieval. The proposed JTANet is a joint venture of the autoencoder and siamese networks. The encoder network of autoencoder is shared with the siamese network. The main aim of the proposed model is to learn more discriminative, robust and efficient feature embeddings for the retrieval task. In order to achieve it, three losses, namely autoencoder loss, siamese loss and feature regularization loss are used. The siamese loss is computed from the triplets which is generated from the embeddings itself using Hard, Random Hard and Semi Hard strategies. The image retrieval experiments are conducted over histopathological colon cancer nuclei dataset. The experimental results suggest that the Semi Hard triplet selection method with 512 embedding length is the most suitable for JTANet with 1:5:1 weighting between autoencoder, siamese and regularization losses. It is also observed that the proposed JTANet model exhibits the faster convergence property. The autoencoder loss is not important with Hard triplet selection strategy. It is also noticed that the siamese loss dominates over other losses in Random Hard and Semi Hard strategies. The experimental results confirm the suitability of the proposed JTANet model for RCC patch retrieval.

\section{Acknowledgment}
This research is funded by Science and Engineering Research Board (SERB), Govt. of India through Project Sanction Number ECR/2017/000082. The authors would like to thank NVIDIA Corporation for the support of 2 GeForce Titan X Pascal GPUs.

\bibliographystyle{elsarticle-num}
\bibliography{References}
\end{document}